\documentclass[letterpaper, 10 pt, conference]{ieeeconf}  % Comment this line out
\IEEEoverridecommandlockouts                              % This command is only
\usepackage{graphics} % for pdf, bitmapped graphics files
\usepackage{amsmath} % assumes amsmath package installed
\usepackage{amssymb}  % assumes amsmath package installed
\usepackage{graphicx}
\title{\LARGE \bf
Image Processing Techniques \\ for Indoor and Outdoor Self Driving cars
}

\author{Rohit Gandikota$^{1}$% <-this % stops a space
\thanks{*This work was done as a course project for AV489 Image and Video Processing. Also this is a start for the IIST's autonomous car for mail delivery project}% <-this % stops a space
\thanks{$^{1}$Rohit Gandikota is an Undergrad from Department of Avionics, Electrical and Communication Engineering,
        Indian Institute of Space science and Technology, Valiyamala, Kerala, India
        {\tt\small grohit0 at gmail dot com}}%
}

\begin{document}

\maketitle
\thispagestyle{empty}
\pagestyle{empty}

%%%%%%%%%%%%%%%%%%%%%%%%%%%%%%%%%%%%%%%%%%%%%%%%%%%%%%%%%%%%%%%%%%%%%%%%%%%%%%%%
\begin{abstract}
In this work we try to implement Image Processing techniques in the area of autonomous vehicles, both indoor and outdoor. The challenges for the both are different and the ways to tackle them varies too. We also showed deep learning makes things easier and precise. We also made base models for all the problems we tackle while building an autonomous car for IIST. 
\end{abstract}

%%%%%%%%%%%%%%%%%%%%%%%%%%%%%%%%%%%%%%%%%%%%%%%%%%%%%%%%%%%%%%%%%%%%%%%%%%%%%%%%
\section{INTRODUCTION}

Machine Learning and Artificial Intelligence has made human lives so at ease, that we wish to automate everything now. From bread toaster to schooling system everything is automated. Similarly driving has caught the attention of machine learning scientists. We have tried to tackle with two sorts of environment for the vehicle.
\begin{itemize}
    \item Indoor environment 
    \item Outdoor environment
\end{itemize}
Both the methods are similar in few ways, but have their own set of challenges and concerns. So in this section we would go on to introduce the same for both the environments in detail.
\subsection{Indoor Autonomous Vehicle}
This might be used as a rover inside a lab, or an autonomous cleaner. Sometimes an autonomous teacher too. So the challenges and concerns for this sort of a problem would be
\begin{itemize}
    \item The The Floor and it's stains and patterns
    \item Obstacle Detection. 
    \item Map formation
    \item Self Localization
\end{itemize}
A robot should avoid an obstacle and travel on the floor. So identifying the obstacle and floor is the first problem. We can solve this by segmentation techniques. Now usually segmentation uses intensities or gradients. So, if our floor is stained or has some different pattern from what we trained it on, can be detected as obstacle. This was simply solved by blurring the image just a little. Recognizing an obstacle is a short term goal for an autonomous vehicle. It should be able to memorize the map of the environment so that it can navigate through the path freely. Now this feature can be taken off from our navigation robot and make it short sighted too, but when you have a specific destination to reach, map formation might help the robot to reach fast and in an efficient way. Map formation method opted in this work is explained in detail in the next section.
\subsection{Outdoor Autonomous Vehicle}
This can be used as a cab, mail delivery vehicle or even outdoor scavenger. The challenges and concerns for this sort would be different than the indoor one. Since the driving speed and size are going to be faster and bigger, we considered the following problems
\begin{itemize}
    \item Basic Lane detection
    \item Road Sign Detection and Identification
    \item Steering Angle Prediction 
    \item Vehicle Detection and Proximity Determination
\end{itemize}
This car should avoid hitting a fellow car or human to full extent. Also we should make sure it adapts it's behaviour based on the road. This can be estimated by using the Road signs. Also giving it a brain to drive can be seen as a regression problem in Deep learning by predicting steering angle and throttle.  We tackled this sophisticated problem using a CNN. Now finally detecting a fellow vehicle and determining a proximity around it where we can not drive into is very important for the safety of our vehicle and the humans. Therefore vehicle detection was tackled using HOG features fed to train SVM as classifier. In the further sections we explain in detail all these aforementioned problems and finally conclude by talking about future work and how we can integrate deep learning and image processing into an Autonomous Vehicle.
\section{Indoor Autonomous Vehicle}
In this section we explain in detail, the steps we have taken to build the vision part of an indoor robot. Also we talk about the challenges aforementioned in detail. \par
We started out with the camera calibration. Camera calibration refers to calibrating our camera to transform the pixels to length in real world. We did not require any transformation from camera co-ordinates to real-world co-ordinates as the method we took for self-localization was also using image processing. This reduced our work a lot. \par
	So we made a set-up using a book of known dimensions, a measuring tape and the camera to be calibrated. We installed the book to in a plain white background and clicked the photos of the book from known locations. This was done by using measuring tape. This was the data collection part. \par 

\begin{figure}[thpb]
      \centering
       \includegraphics[scale=0.5]{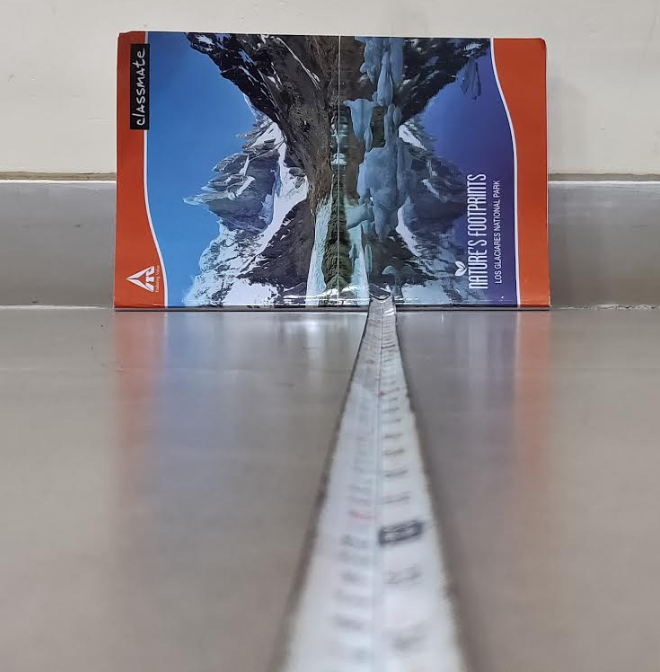}
      \caption{A photo taken from 70cm of a book of known dimensions for camera calibration}      
        \label{figurelabel}
   \end{figure}
   Now we went on to process the image using a python tool, openCV. We smoothened the image using Gaussian filter of size 3x3 after converting the image to greyscale. Later we applied sobel operator of size 3x3 to the image and thresholded the the edges to obtain strong edges. Then using hough transform we detected rectangles, as the book is rectangle in shape. Finally we obtained the rectangle of highest area. This was done using the function \(cv2.findContours\) and \(cv2.contourArea\). Now we find the number of pixels \(N\) covering the book’s edge whose known length is \(L\) taken from distance \(D\). We apply a formula to find the perceived focal length \(F\)
   $$
     F = (N * D)/L
   $$
   Later this can be used to obtain the distance \(D1\) to the objects in real world from the image captured by the calibrated camera using this formula. Say their length in the captures image (number of pixels is \(P\). 
   $$
    D1 = (W * F) / P
   $$
   Now once the image is captured our first goal is to recognize the obstacles. So this can be seen as a two class segmentation problem by a Computer Vision guy. We have used 3 methods to segment the floor from obstacles. We compare the results and select one for the final version. \par 
\subsection{Watershed Segmentation}
Watershed is a transformation defined on a grayscale image. The name refers metaphorically to a geological watershed, or drainage divide, which separates adjacent drainage basins. The watershed transformation treats the image it operates upon like a topographic map, with the brightness of each point representing its height, and finds the lines that run along the tops of ridges.
The results for watershed segmentation are shown in Figure.2.
\begin{figure}[thpb]
      \centering
       \includegraphics[scale=0.8]{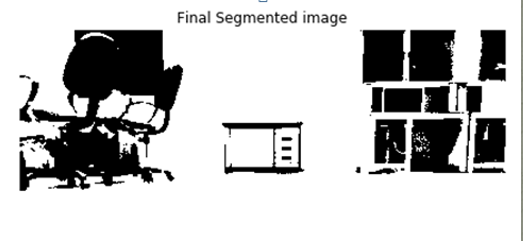}
      \caption{Result of watershed segmentation after preprocessing. Black representing obstacles.}      
        \label{figurelabel}
   \end{figure}
\subsection{K-means segmentation}
K-Means is a least-squares partitioning method that divide a collection of objects into K groups. The algorithm iterates over two steps:
\begin{itemize}
    \item Compute the mean of each cluster.
    \item Compute the distance of each point from each cluster by computing its distance from the corresponding cluster mean. Assign each point to the cluster it is nearest to.
\end{itemize}
Iterate over the above two steps till the sum of squared within group errors cannot be lowered any more. The results for K-mean clustering were shown in Figure.3.

\begin{figure}[thpb]
      \centering
       \includegraphics[scale=0.7]{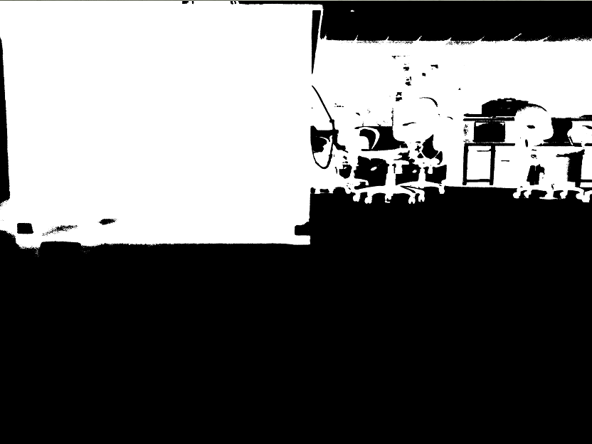}
      \caption{Results of K-mean segmentation after preprocessing. White representing obstacles.}      
        \label{figurelabel}
   \end{figure}
\subsection{Otsu thresholding}
The algorithm assumes that the image contains two classes of pixels following bi-modal histogram (foreground pixels and background pixels), it then calculates the optimum threshold separating the two classes so that their combined spread (intra-class variance) is minimal, or equivalently (because the sum of pairwise squared distances is constant), so that their inter-class variance is maximal. This is a hard-search 2 class problem, for all the thresholds we try to find a threshold which maximizes the inter class variance. The results for this algorithm are shown in Figure.4.

\begin{figure}[thpb]
      \centering
       \includegraphics[scale=0.7]{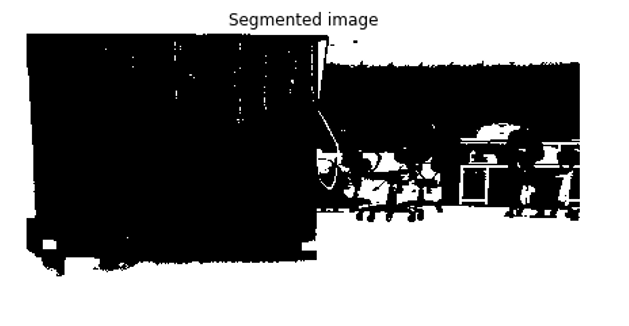}
      \caption{Results of Otsu-segmentation after preprocessing. Black representing obstacles.}      
        \label{figurelabel}
   \end{figure}
\textbf{Map Formation:}
The later step in this work was to make the robot learn the environment. This was done using Image stitching. Every time the robot finds the obstacles and free path, we recomputed a path and give instructions to the motor. Based on the movement we calculate the new orientation using normal rotation and translation matrix. The origin of the map is set during the starting of the robot’s exploration part. Now that we have the orientation of the robot, we add the segmented image to the map by just patching the new image to the map image. This is show in Figure.5.
\begin{figure}[thpb]
      \centering
       \includegraphics[scale=0.7]{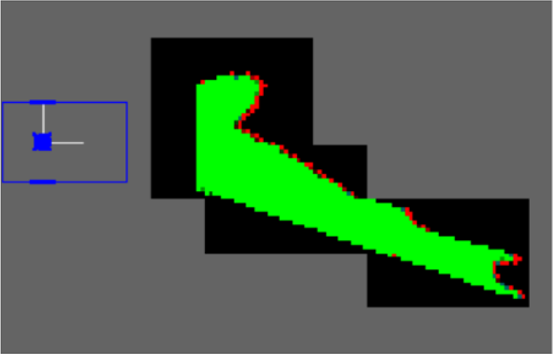}
      \caption{Map formation of CVPR Lab of IIST.}      
        \label{figurelabel}
   \end{figure}
   \begin{itemize}
       \item At start the robot has no information about the environment. To create the map the start position of the robot is defined, and a destination assigned. 
       \item To reach the destination the robot will explore the world, looking at the floor and generating the first obstacles. 
       \item Every time a new obstacle is detected, the robot computes the path to reach the destination, and the map grows.
       \item  To map the entire environment it will be enough to give a goal location unreachable, as outside a wall.
   \end{itemize}
  Finally after the robot has learned the map of an environment, it can self-localize itself by correlating some features in the image taken in the moment with the learned map. The comparison is based on the angles between the walls. The robot can upgrade its location matching the global and the local maps. The self-location problem is important when the robot has to move in autonomous way. Dead reckoning reduces the location error, but is unable to keep the error under a given threshold.  Robot starts from any position, builds a map with any origin, but we need to match the new map with the standard one that is in the user interface. So the problem is to match a partial map onto a complete map. This method is depicted in Figure.6.
  \begin{figure}[thpb]
      \centering
       \includegraphics[scale=0.4]{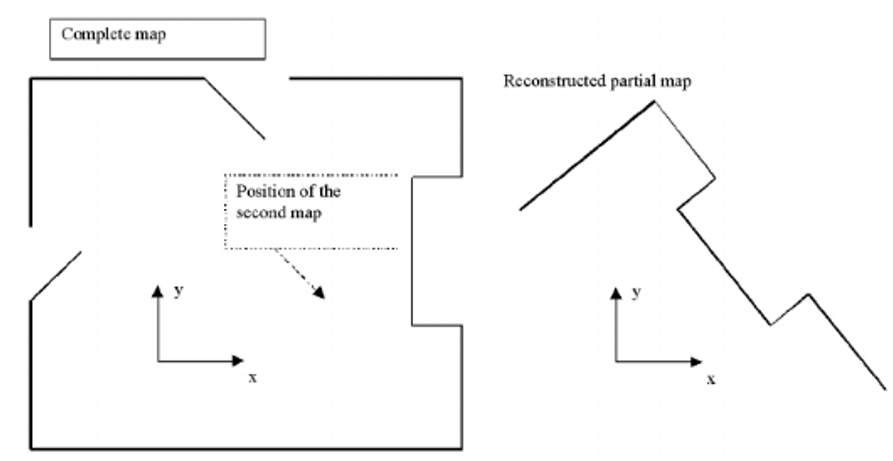}
      \caption{Method to self-localize in a learned Map.}      
        \label{figurelabel}
   \end{figure}

\section{Outdoor Autonomous Vehicle}
In this section we explain in detail the work we have done to make some base models and lay a foundation for IIST's autonomous vehicle for mail delivery. For this work we mostly used Machine learning and Deep Learning. Although we have used a lot of Image Processing during pre-processing or for feature extraction. The following tasks have been covered.
\begin{itemize}
    \item Lane Detection
    \item Road sign Identification (Deep Learning)
    \item Steering Angle Prediction (Deep Learning) 
    \item Vehicle Detection
\end{itemize}
The dataset is open sourced by Udacity’s “self-driving car” course. But we would like to collect data from IIST to best train the car. Further we explain all the tasks in detail. 

\subsection{Lane Detection}
Late detection refers to identifying the lane we are moving on. This was done using image processing techniques like smoothing followed by Edge detection and thresholding. We finally detect the lanes using hough transform. The steps are show in a image sequence below. Also a code snippet is shown in Fig.9.
  \begin{figure}[thpb]
      \centering
       \includegraphics[scale=0.35]{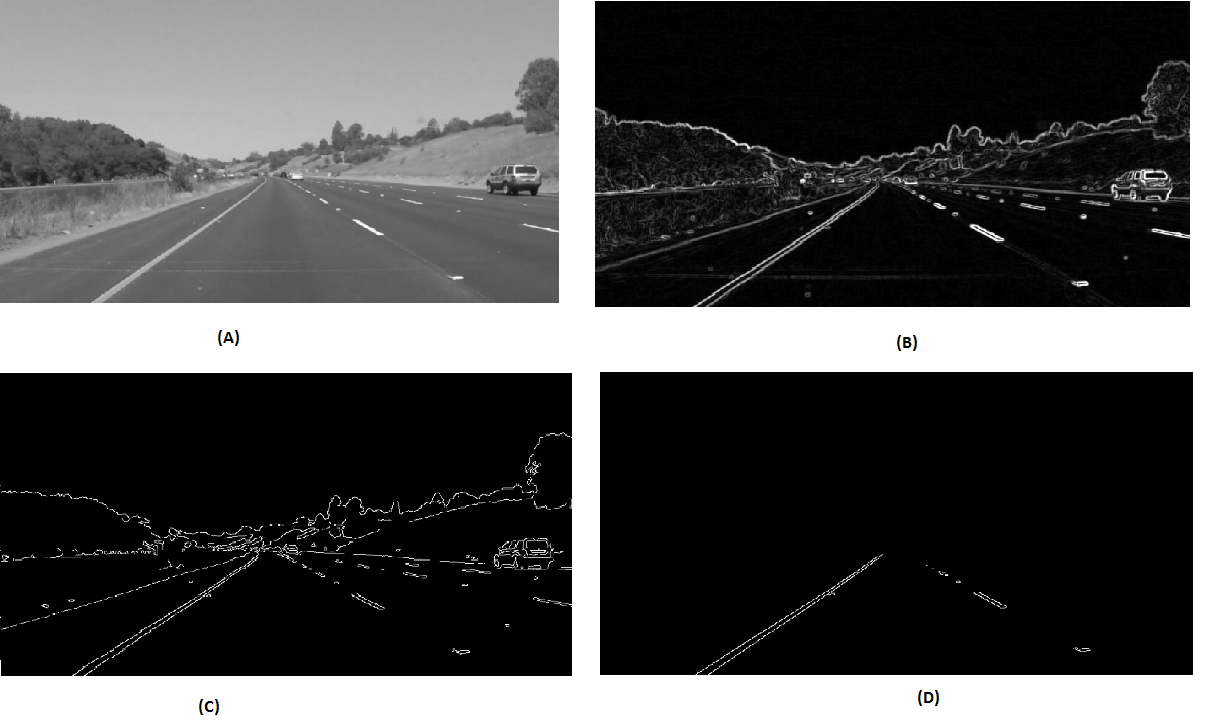}
      \caption{Sequence followed to tackle the lane detection problem.}      
        \label{figurelabel}
   \end{figure}
In Fig.7.(A), we have converted the image from RGB to grayscale and applyed a smoothing filter (Gaussian). In Fig.7.B we have applied a sobel operator and then thresholded this image to give strong edges as shown in Fig.7.c. Finally we applied a mask to keep the lane in the centre of the image as that is where our lane is present.( Assuming that the camera is mounted on the top centre of the car). Finally we do hough transform to get two lines which will be our lane. The Final output is shown in Fig.8. This method can be applied to videos as well. 
  \begin{figure}[thpb]
      \centering
       \includegraphics[scale=0.4]{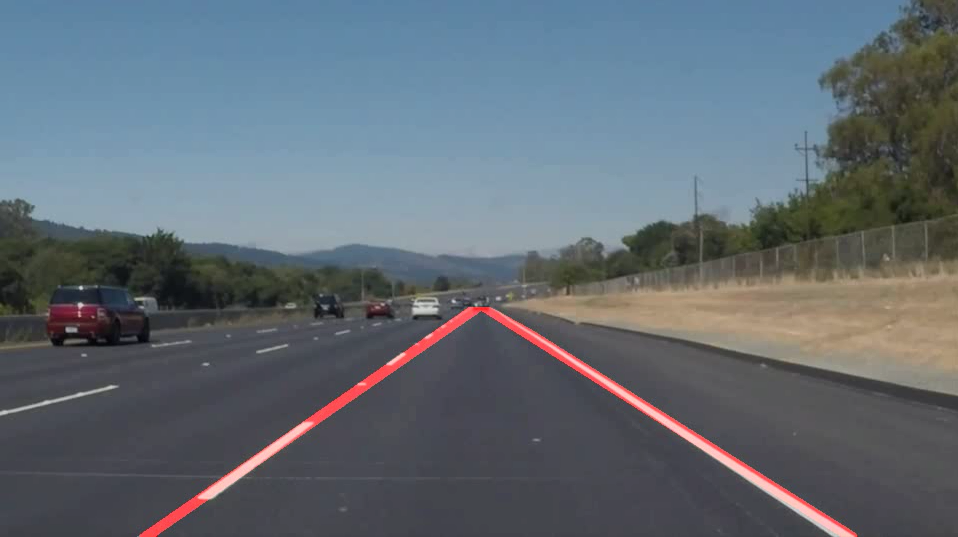}
      \caption{The final output for the lane detection. The red lines form the detected Lane.}      
        \label{figurelabel}
   \end{figure}
   \begin{figure}[thpb]
      \centering
       \includegraphics[scale=0.6]{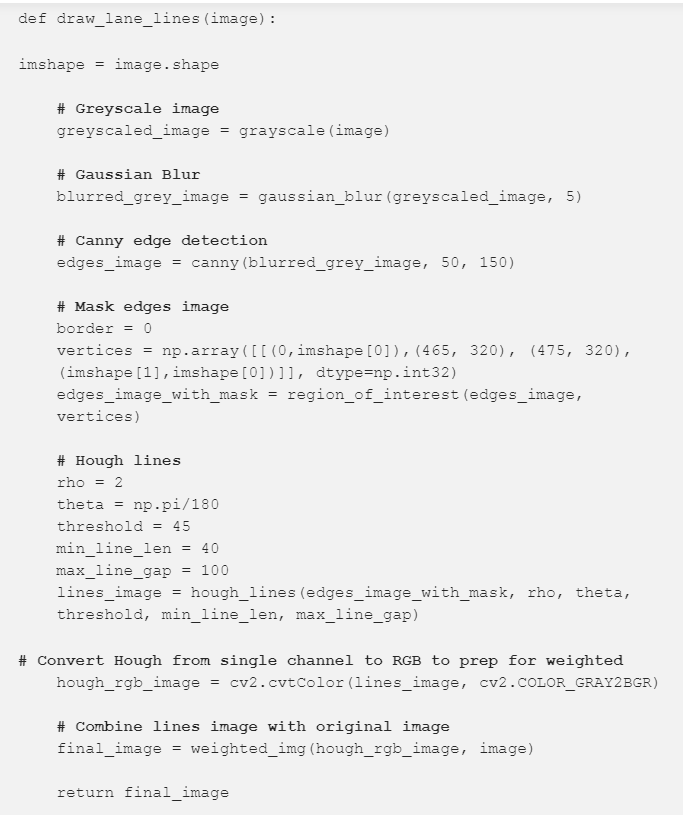}
      \caption{Pseudo-code for Lane Detection.}      
        \label{figurelabel}
   \end{figure}
\subsection{Road Sign Identification}
Road sign identification was done using a CNN. The dataset was open-sourced German Traffic Dataset. This was a simple straight forward problem. We faced a problem of under fitting although we tackled it by data augmentation. We used batch normalization and dropout of 0.5 for training. This gave a training accuracy of 99\%, validation accuracy of 98\% and a test accuracy of 98\%. The architecture used had 3 Convolution and pooling layers stacked on 1 fully connected and finally output. We are not going to go into implementation details as this is a basic problem. The test results are shown below. 
   \begin{figure}[thpb]
      \centering
       \includegraphics[scale=0.8]{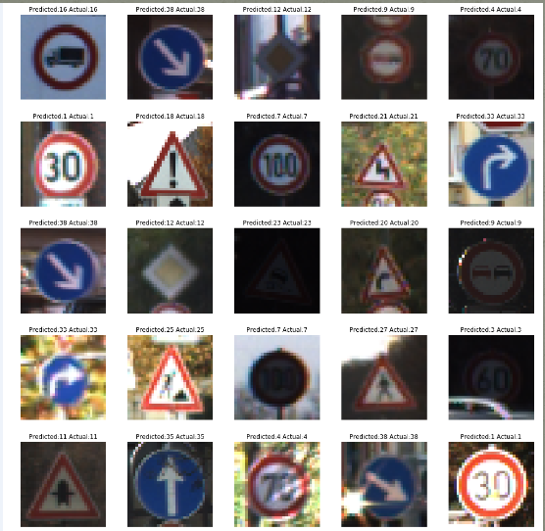}
      \caption{Road sign Identification CNN's test results showing 100\% true positives .}      
        \label{figurelabel}
   \end{figure}

\subsection{Steering Angle Prediction}
Here the problem statement is to predict steering angle based on the video feed from the camera mounted on our autonomous car. This was tackled by using a CNN. Although this is just a preliminary attempt, we wish to further work on LSTM inside CNN model. The following problems were faced during the training of CNN. \begin{itemize}
    \item CNN not able to regress so nice
    \item Even after solving the above problem, CNN giving sudden jerky angles per frame
    \item And finally not learning enough (Underfitting)
\end{itemize}

We have solved the above mentioned problems in the following way. The first problem was solved by binning the angles. Binning refers to giving a single output for angle ranges. The second problem was solved by upgrading the loss function. We have added a smooth regularizer on consecutive frame outputs. This helped in smoothing the steering a lot. Now we solved this under-fitting by augmenting data. Data augmentation involved flipping and rotation. Also blurring and motion distortion. This helped in robust steering. We finally visualized the output by adding an angle rotated steering wheel photo. A snapshot is shown in Fig. 11.

   \begin{figure}[thpb]
      \centering
       \includegraphics[scale=0.48]{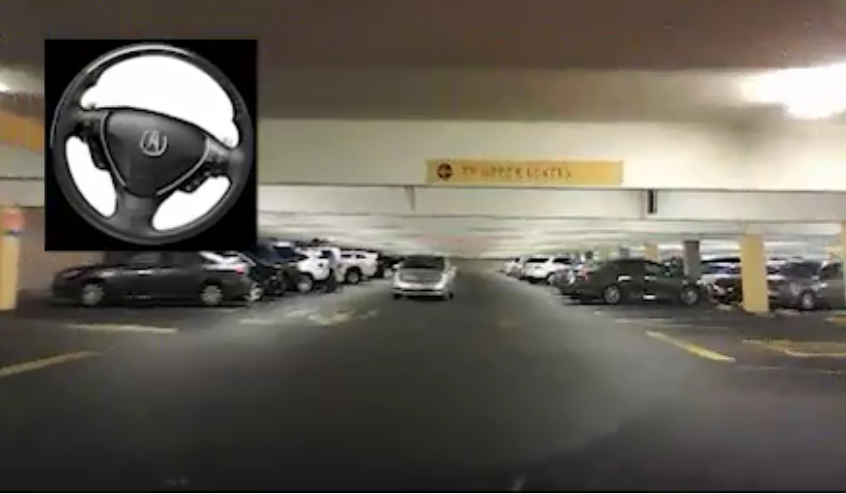}
      \caption{Output of a steering angle prediction being visualized in a video. This is just a snapshot of a video.}      
        \label{figurelabel}
   \end{figure}
   The network architecture used in this work is shown in fig.12
      \begin{figure}[thpb]
      \centering
       \includegraphics[scale=0.7]{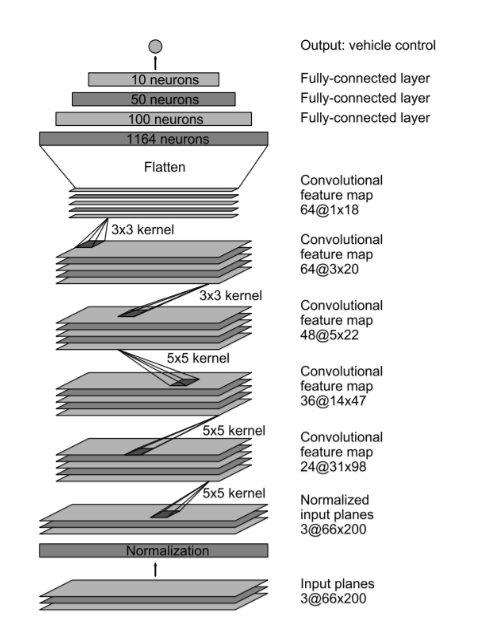}
      \caption{The network architecture used for vehicle detection. As can be seen this was not time series model. This will be augmented in future.}      
        \label{figurelabel}
   \end{figure}
\subsection{Vehicle Detection and Proximity Detection for Collision Avoidance}
The goal of the project was to develop a pipeline to reliably detect cars given a video from a roof-mounted camera: in this section the reader will find a short summary of how we tackled the problem. In the field of computer vision, a features is a compact representation that encodes information that is relevant for a given task. In our case, features must be informative enough to distinguish between car and non-car image patches as accurately as possible.
Fig. shows how the vehicle and non-vehicle classes look like in this dataset
      \begin{figure}[thpb]
      \centering
       \includegraphics[scale=0.52]{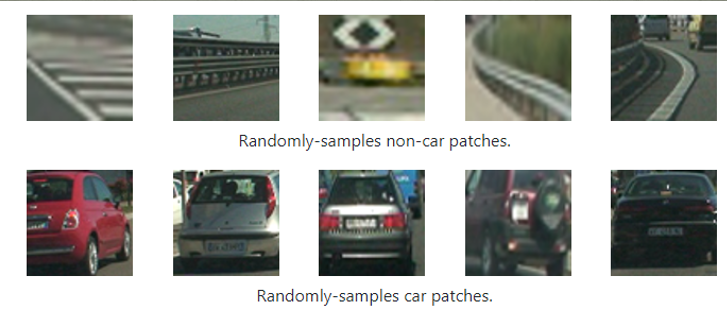}
      \caption{Random samples of cars and non-cars from the dataset by Udacity.}      
        \label{figurelabel}
   \end{figure}
For the task of car detection We used color histograms and spatial features to encode the object visual appearance and HOG features to encode the object's shape. While color the first two features are easy to understand and implement, HOG features can be a little bit trickier to master. Choosing the parameters for HOG were really hard, we chose some finally by doing trail and error. HOG stands for Histogram of Oriented Gradients and refer to a powerful descriptor that has met with a wide success in the computer vision community, since its introduction in 2005 with the main purpose of people detection. 
      \begin{figure}[thpb]
      \centering
       \includegraphics[scale=0.52]{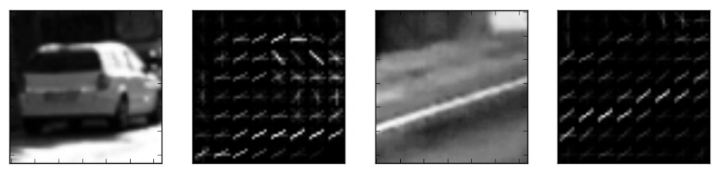}
      \caption{Representation of HOG descriptors for a car patch (left) and a non-car patch (right).}      
        \label{figurelabel}
   \end{figure}
   The bad news is, HOG come along with a lot of parameters to tune in order to work properly. The main parameters are the size of the cell in which the gradients are accumulated, as well as the number of orientations used to discretize the histogram of gradients. Furthermore, one must specify the number of cells that compose a block, on which later a feature normalization will be performed. Finally, being the HOG computed on a single-channel image, arises the need of deciding which channel to use, eventually computing the feature on all channels then concatenating the result.
In order to select the right parameters, both the classifier accuracy and computational efficiency are to consider.\par
Once decided which features to used, we can train a classifier on these.We train a linear SVM for task of binary classification car vs non-car. SVM is said to have very high false positives. Now in an Image how do w detect a car. We opted a method which combines HOG feature extraction with a sliding window search, but rather than perform feature extraction on each window individually which can be time consuming, the HOG features are extracted for the entire image (or a selected portion of it) and then these full-image features are sub-sampled according to the size of the window and then fed to the classifier. The method performs the classifier prediction on the HOG features for each window region and returns a list of rectangle objects corresponding to the windows that generated a positive ("car") prediction. The image below shows the first attempt at using this method on one of the test images, using a single window size. 
\begin{figure}[thpb]
      \centering
       \includegraphics[scale=0.6]{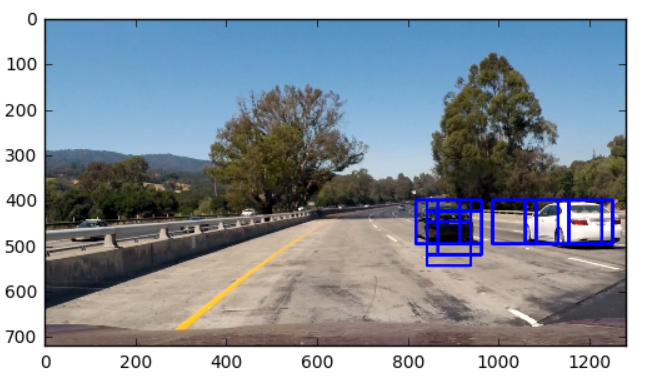}
      \caption{Result of Image detection using a single window size.}      
        \label{figurelabel}
   \end{figure}
Now we used a simple trick of how objects look smaller as they go far. So we used a set of window sizes for each distance from camera. This is shown in Fig.16. They sum up to be 697 windows in total. To reduce computational burden we used the same trick as mentioned above.
\begin{figure}[thpb]
      \centering
       \includegraphics[scale=0.4]{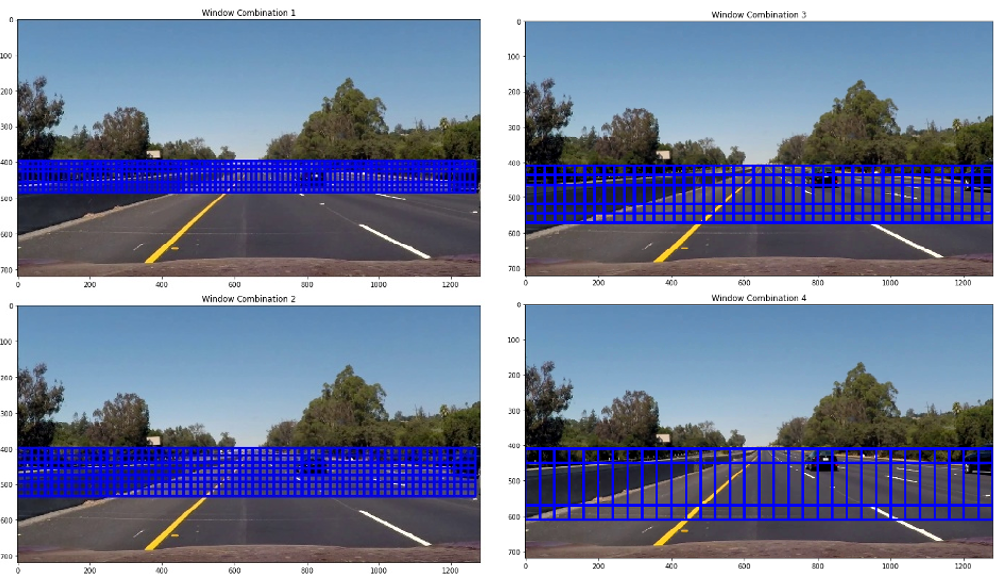}
      \caption{Image showing different sliding window sizes for different distances from the camera. }      
        \label{figurelabel}
   \end{figure}
Now detecting car is giving a best fit for the bounding box. Also we see multiple boxes around the car along with some false positives. This problem was solved using heatmapping and thresholding. We drew gaussians around the centre point of each boxes and thresholded them with 0.5*max. This flow is shown below.
\begin{figure}[thpb]
      \centering
       \includegraphics[scale=0.5]{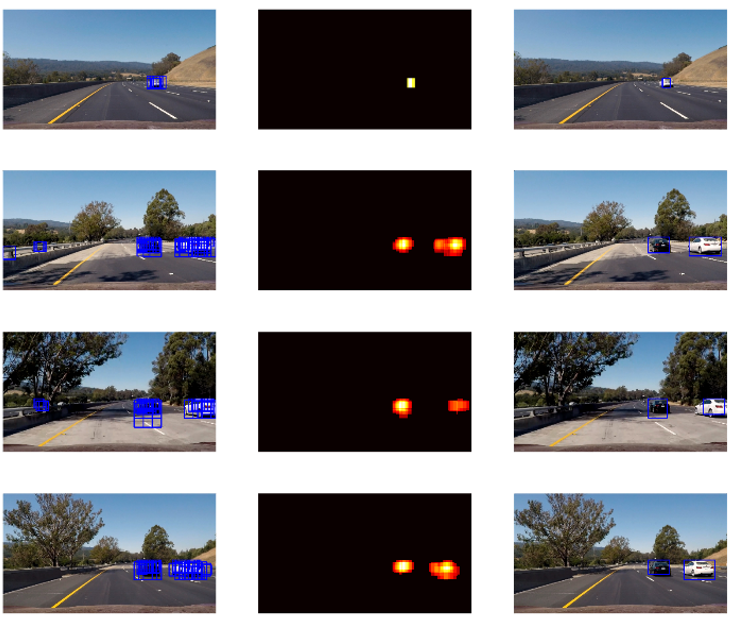}
      \caption{Left images show the pre-filtered boxes. Middle images show the heatmaps of the boxes. And the right images show a single bounding box around the car and no false positives. }      
        \label{figurelabel}
   \end{figure}
We have integrated all the above mentioned Techniques for the Outdoor Autonomous robot and the result is shown in Fig.18. 
\begin{figure}[thpb]
      \centering
       \includegraphics[scale=0.5]{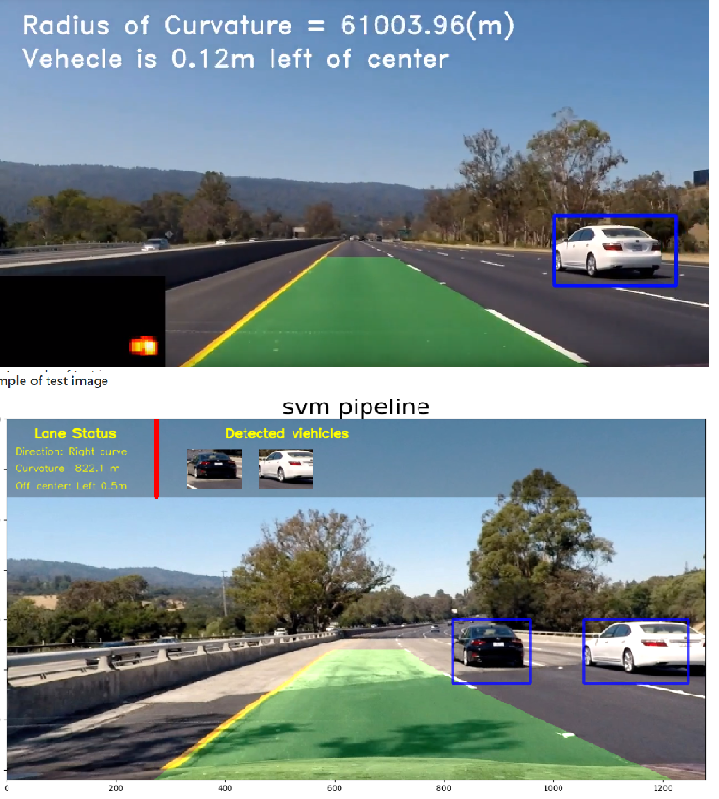}
      \caption{Integration of all the processes we discussed separately.}      
        \label{figurelabel}
   \end{figure}

\section{CONCLUSIONS}

We conclude by saying we have achieved to build a solid base for IIST's autonomous mail delivering car. We have discussed many aspects one should consider while building Vision to autonomous robots. We have also explained in detail how to deal with certian problems while dealing with Computer Vision problems. We would like to further work on steering angle prediction by getting time series dependencies into play. We wish to do it by building a CNN with LSTMS inbuilt. Later we would like to use Single Shot Detection to Detect all the objects like cars, humans and road signs. We would also like to get more Deeplearning into the picture more.   

\addtolength{\textheight}{-12cm}   % This command serves to balance the column lengths
                                  % on the last page of the document manually. It shortens
                                  % the textheight of the last page by a suitable amount.
                                  % This command does not take effect until the next page
                                  % so it should come on the page before the last. Make
                                  % sure that you do not shorten the textheight too much.

%%%%%%%%%%%%%%%%%%%%%%%%%%%%%%%%%%%%%%%%%%%%%%%%%%%%%%%%%%%%%%%%%%%%%%%%%%%%%%%%

%%%%%%%%%%%%%%%%%%%%%%%%%%%%%%%%%%%%%%%%%%%%%%%%%%%%%%%%%%%%%%%%%%%%%%%%%%%%%%%%

%%%%%%%%%%%%%%%%%%%%%%%%%%%%%%%%%%%%%%%%%%%%%%%%%%%%%%%%%%%%%%%%%%%%%%%%%%%%%%%%

\newpage
\section*{ACKNOWLEDGMENT}
Prof. Deepak Mishra who is with faculty of Avionics has guided us in this work. Kaninika Pant, a fellow undergrad, has also contributed to this work with her immense support both technically and morally. Most of the work is referred from online blogs and dataset credits to Udacity.

%%%%%%%%%%%%%%%%%%%%%%%%%%%%%%%%%%%%%%%%%%%%%%%%%%%%%%%%%%%%%%%%%%%%%%%%%%%%%%%%

\end{document}